\PassOptionsToPackage{table}{xcolor}
\PassOptionsToPackage{normalem}{ulem}

\documentclass[]{bytedance_seed}

\usepackage[toc,page,header]{appendix}
\usepackage{minitoc}

\usepackage{amsmath}
\usepackage{amsfonts}
\usepackage{nicefrac}
\usepackage{float}
\usepackage{algorithm}
\usepackage{listings}
\usepackage{marvosym} % for \Letter (envelope symbol for corresponding author)
\newcommand{\corrmark}{\text{\Letter}}
\title{Representation Forcing for Bottleneck-Free Unified Multimodal Models}

\author[1,2\,\dagger]{Yuqing Wang}
\author[2\,\ddagger]{Zhijie Lin}
\author[2]{Ceyuan Yang}
\author[2]{Yang Zhao}
\author[2]{Fei Xiao}
\author[3,2\,\dagger]{Hao He}
\author[2]{\\Qi Zhao}
\author[2]{Zihan Ding}
\author[3,2\,\dagger]{Fuyun Wang}
\author[4,2\,\dagger]{Shuai Wang}
\author[5,2\,\dagger]{Youliang Zhang}
\author[2]{\\Haoqi Fan}
\author[1\,\corrmark]{Xihui Liu}

\affiliation[1]{University of Hong Kong}
\affiliation[2]{ByteDance Seed}
\affiliation[3]{The Chinese University of Hong Kong}
\affiliation[4]{Nanjing University}
\affiliation[5]{Tsinghua University}

\renewcommand{\affiliationlist}{%
  \affiliationformat[1]{University of Hong Kong},\;%
  \affiliationformat[2]{ByteDance Seed}\\%
  \affiliationformat[3]{The Chinese University of Hong Kong},\;%
  \affiliationformat[4]{Nanjing University},\;%
  \affiliationformat[5]{Tsinghua University}%
}

\abstract{
Unified multimodal models (UMMs) aim to handle perception and generation in a single model. Yet existing UMMs still rely on a frozen, separately pretrained VAE for image generation, imposing a structural bottleneck. Naively removing it introduces a quality gap, as the model must learn both high-level structure and low-level details from raw pixels. In this paper, we propose Representation Forcing (RF), a technique that closes this gap by making representation prediction a native capability of the model. Concretely, RF forces the decoder to autoregressively predict visual representations as intermediate tokens before pixels; these tokens then stay in context to guide pixel diffusion within the same backbone. By turning representations from perception outputs into generation targets, RF eliminates the need for any external generative latent space. We find that RF benefits both understanding and generation. On image generation, our pixel-space model with RF matches state-of-the-art VAE-based unified models. On image understanding, pixel-space RF generally outperforms its VAE-based variant. Together, these results offer an effective step toward end-to-end, bottleneck-free UMMs.
}

\date{June 1, 2026}
\checkdata[Project Page]{\url{https://yuqingwang1029.github.io/RepresentationForcing}}

\begin{document}
\maketitle

% Footnotes at the bottom of the first page (custom marks match the author superscripts).
{\renewcommand{\thefootnote}{$\dagger$}%
 \footnotetext{Work done during an internship at ByteDance Seed.}}
{\renewcommand{\thefootnote}{$\ddagger$}%
 \footnotetext{Project lead.}}
{\renewcommand{\thefootnote}{\corrmark}%
 \footnotetext{Corresponding author.}}
\setcounter{footnote}{0}

% 不需要目录就注释掉 注意目录不要和第一页放在一块 要有\newpage
%\newpage
%\tableofcontents
%\newpage

\section{Introduction}
\label{sec:intro}

The ability to perform understanding (text output) and generation (pixel output) within a unified framework represents a fundamental step toward general-purpose multimodal intelligence~\cite{chameleon,emu3,bagel,showo,janus}. Prevailing unified multimodal models (UMMs) pursue this by bringing language and image generation into a shared transformer backbone~\cite{transfusion, janusflow, bagel}, with next-token prediction for language and diffusion for image generation. However, despite this unification, the image generation pathway still depends on a separately pretrained, frozen VAE~\cite{vae,esser2020taming,stable_diff}: images are compressed into latents before diffusion is applied, and pixels are recovered through a fixed decoder. This creates a structural bottleneck. The latent space is optimized for reconstruction rather than the objectives of the unified model, and its lossy compression imposes a hard upper bound on generation quality that further training of the UMM cannot overcome. Removing this bottleneck is an important step toward end-to-end UMMs.

\begin{figure}[t]
\centering
\includegraphics[width=\linewidth]{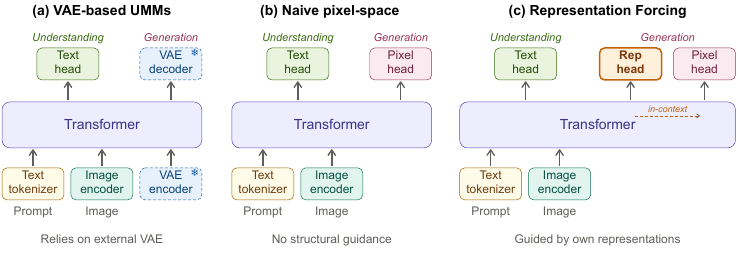}
\caption{\textbf{Architectural comparison.} (a)~Prevailing UMMs rely on a frozen VAE encoder and decoder for image generation, creating a structural bottleneck. (b)~Naively removing the VAE and generating directly in pixel space eliminates this bottleneck but loses structural guidance, leading to a quality gap. (c)~Representation Forcing closes this gap by training the transformer decoder to autoregressively predict visual representations (Rep head) before pixel generation. These representations are trained to match features from the model's own understanding encoder and remain in context within the shared transformer, providing structural guidance for pixel-space diffusion without any external latent space.}
\label{fig:teaser}
\end{figure}

A natural alternative is to generate directly in pixel space. Recent works have shown pixel-space diffusion to be feasible for standalone generation models~\cite{jit,pixelflow,pixnerd}. However, we find that directly applying these methods in UMMs fails to match the quality of VAE-based counterparts. 
We attribute this to the broader image distribution and richer text conditioning in UMMs: the model must learn both the high-level semantic structure and fine-grained details of an image from the same raw signal. 
This motivates an intermediate representation that separates these two factors, so that the diffusion process can focus on low-level rendering without falling back to an external latent space.
% We attribute this to the lack of a visually-grounded intermediate abstraction between text and raw pixels: the model must learn both the high-level semantic structure and fine-grained details of an image from the same raw signal. 

The key question is where such a representation should come from. We observe that UMMs already provide one internally: their understanding pathway learns visual representations that capture high-level structure, such as object identity, spatial layout, and scene composition. In understanding, the encoder~\cite{siglip,tschannen2025siglip2,dinov2,dinov3} extracts these representations from observed images. In generation, however, no image is available, and the model must produce them from the input context alone. This means the model must learn to predict these representations on its own.

In this paper, we propose \textbf{Representation Forcing} (RF), an approach that closes this gap by making representation prediction a native capability of the model. Our key idea is to ground the high-level visual representation in the decoder itself. Concretely, we use visual representations extracted by the understanding encoder as targets, and train the model decoder to predict them autoregressively under the same next-token prediction objective used for language. These predicted representations provide an explicit structural scaffold between text and pixels; they remain in the sequence as in-context conditioning, guiding pixel-space generation within the shared transformer. By turning representations from perception outputs into generation targets, RF grounds understanding and generation in a single representation space, without relying on a separately pretrained latent space (Figure~\ref{fig:teaser}).

\begin{figure}[t]
\centering
% \vspace{-5pt}
\includegraphics[width=\linewidth]{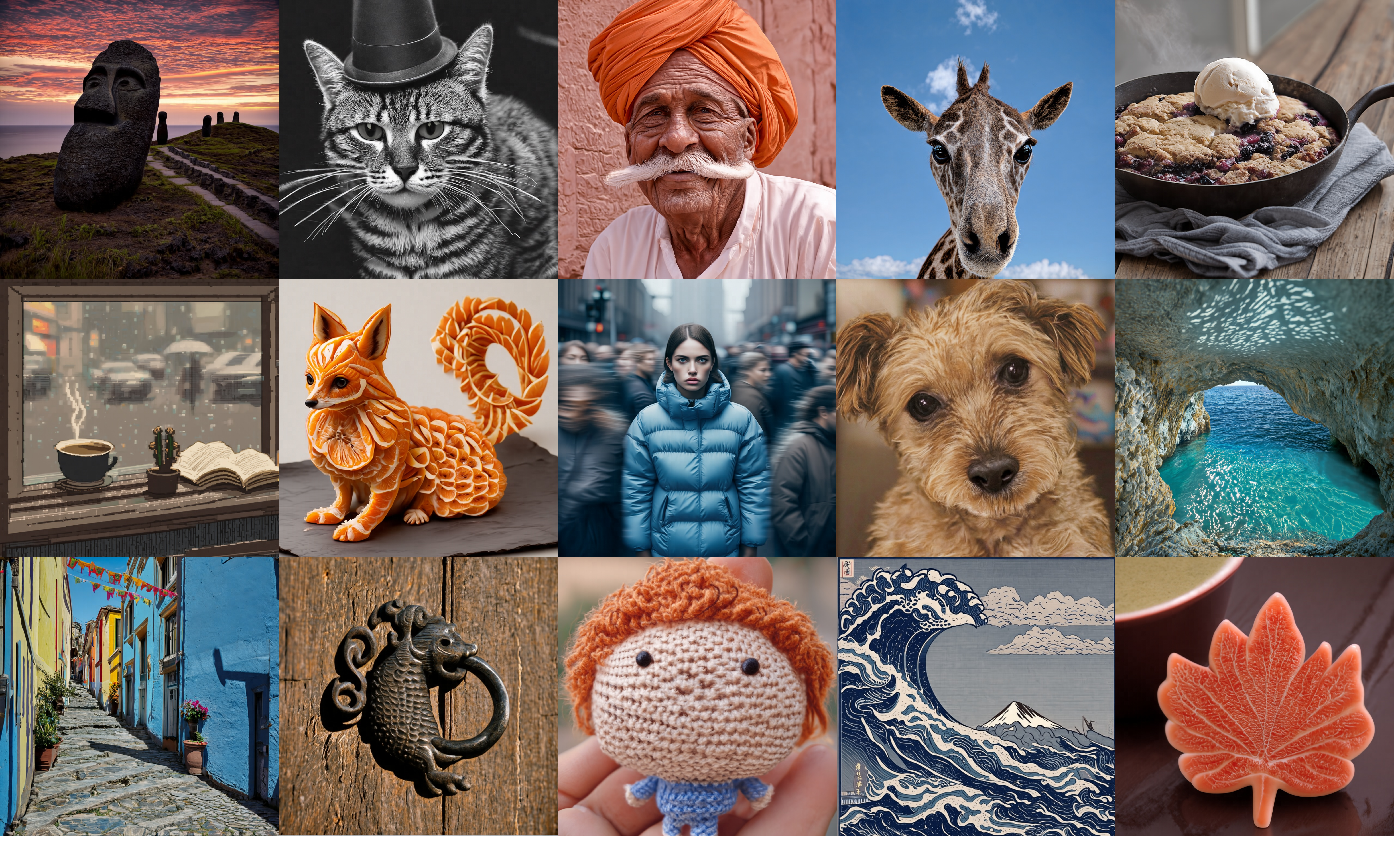}
\caption{Text-to-image generation results at $1024 \times 1024$ resolution from our pixel-space unified model with Representation Forcing.}
\label{fig:demo}
\end{figure}

To validate Representation Forcing, we apply it to both pixel-space and VAE-based UMMs under controlled settings with the same architecture, data, and training budget. On image generation, our pixel-space model with RF matches the VAE-based baseline across standard benchmarks while preserving rich textural details (Figure~\ref{fig:demo}). On understanding, pixel-space RF outperforms its VAE-based variant, showing that pixel-space generation is more compatible with unified multimodal modeling than VAE-based generation. Ablations further reveal that RF is critical for pixel-space generation, while also bringing improvements to VAE-based settings. Together, these findings demonstrate that Representation Forcing benefits both directions of unified multimodal modeling, offering an effective step toward pixel-space, bottleneck-free UMMs.

The main contributions of this work are summarized as follows:
\begin{itemize}
\item We propose Representation Forcing (RF), a simple approach that closes the quality gap of pixel-space image generation in unified multimodal models, eliminating the need for any pretrained VAE. RF trains the decoder to autoregressively predict visual representations as intermediate tokens, which then serve as in-context structural guidance for pixel-space diffusion within the same backbone.

\item RF benefits both image generation and understanding across pixel-space and VAE-based UMMs. In particular, our pixel-space model with RF matches the VAE-based counterpart on generation and outperforms it on understanding, suggesting that pixel-space generation is more compatible with unified multimodal modeling than VAE-based generation.

\item Our work advocates for unified multimodal models where perception and generation share a single, end-to-end-learned representation space, rather than coordinate across separately pretrained components such as external VAEs. Representation Forcing is a step toward fully unified multimodal models, where all capabilities are learned end-to-end within the model itself rather than inherited from independently trained components.
\end{itemize}

\section{Related Work}
\label{sec:related}

\textbf{Unified Multimodal Models.}
Existing UMMs broadly fall into two families. The first generates within a \emph{single backbone}, either by modeling images as discrete tokens (Chameleon~\cite{chameleon}, Emu3~\cite{emu3}) or by attaching diffusion in a VAE latent space (Transfusion~\cite{transfusion}, Show-o~\cite{showo}, JanusFlow~\cite{janusflow}). These approaches further address the interference between understanding and generation through decoupled visual encoders in Janus series~\cite{janus,janus-pro,janusflow} or modality-specific experts in BAGEL~\cite{bagel}, but all depend on a separately pretrained visual tokenizer---a VQVAE or a continuous VAE---to define the latent space for generation. The second family \emph{stitches an LLM with an external diffusion model}: the LLM predicts visual representations such as CLIP features, which then condition a separately trained diffusion decoder to render images, as in Emu2~\cite{emu2}, SEED-X~\cite{seedx}, BLIP3-o~\cite{blip3o}, and MetaQueries~\cite{metaquery}. More recently, Omni~\cite{omni} expands the set of natively trained modalities to text, images, video, and 3D geometry, and rolls out intermediate multimodal context within the backbone before decoding. A growing line of work further seeks to remove the pretrained VAE that bottlenecks these models. Concurrent with our work, SenseNova-U1~\cite{sensenova} and Tuna-2~\cite{tuna2} build pixel-space UMMs that discard the visual encoder and generate directly from raw patches.
Representation Forcing shares this goal of a bottleneck-free, end-to-end UMM, but takes a different path. It builds on the same MoT backbone design~\cite{bagel} and retains a jointly trained understanding encoder, turning its representations into intermediate generation targets that serve as in-context conditioning for pixel-space diffusion within the same transformer, yielding a fully end-to-end UMM.

\textbf{Pixel-Space Generation.}
State-of-the-art diffusion models~\cite{stable_diff,flux} typically operate in the latent space of a pretrained VAE~\cite{vae}, which reduces compute and enables high-resolution synthesis but prevents end-to-end training. A recent line of work has explored generating directly in pixel space~\cite{DDPM_paper,dhariwal2021diffusion}, enabling end-to-end learning from raw data: JiT~\cite{jit} demonstrates that plain Vision Transformers with $x$-prediction can generate on raw pixels, and other approaches explore alternative pixel-space architectures~\cite{sid2,pixelflow,pixnerd}. These methods mostly focus on standalone generation on the ImageNet~\cite{imagenet} dataset. In UMMs, the broader image distribution and richer text conditioning leave naive pixel-space diffusion with a clear quality gap; Representation Forcing closes this gap by providing a structural scaffold from the model's own understanding encoder.

\textbf{Representation Learning for Generation.}
Several works explore richer visual representations to improve generation. REPA~\cite{repa} aligns intermediate diffusion features with frozen pretrained representations to accelerate training convergence; RAE and related works~\cite{rae,svg,cubid} goes further by replacing the VAE entirely with frozen pretrained encoders such as DINOv2~\cite{dinov2} and SigLIP~\cite{siglip}, providing a semantically richer latent space. Closely related, Latent Forcing~\cite{latentforcing} produces high-level structure before pixels by jointly diffusing pixels and frozen DINOv2 latents under separate noise schedules. Despite their differences, all these methods still rely on a frozen, separately pretrained representation space. Representation Forcing instead unfreezes this space: the understanding encoder is trained jointly and end-to-end with the rest of the model, and its features are discretized online into representation tokens that the decoder learns to predict autoregressively. High-level visual structure thus becomes a native output learned within the model rather than inherited from a fixed external space, consistent with our goal of a bottleneck-free, end-to-end UMM.

\section{Representation Forcing}
\label{sec:method}

\begin{figure}[t]
\centering
%\vspace{-4pt}
\includegraphics[width=\linewidth]{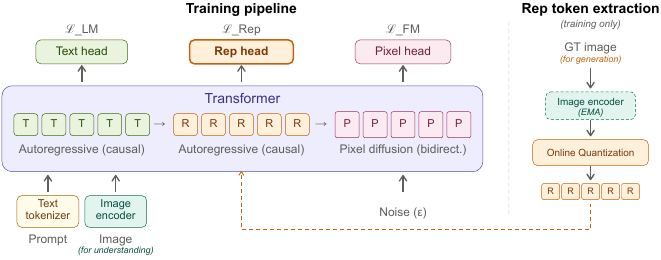}
\caption{\textbf{Training pipeline of Representation Forcing.} \emph{Left:} The decoder processes a unified sequence of text tokens~(T), representation tokens~(R), and pixel patches~(P) within a shared transformer. Text and representation tokens are predicted autoregressively under next-token prediction ($\mathcal{L}_{\mathrm{LM}}$ and $\mathcal{L}_{\mathrm{Rep}}$), while pixel patches are generated via bidirectional diffusion from noise ($\mathcal{L}_{\mathrm{FM}}$). The image encoder provides continuous visual features to the transformer for understanding tasks. \emph{Right:} For generation training, an EMA copy of the image encoder extracts features from the ground-truth image, which are discretized via online quantization into representation tokens. These tokens provide both the training targets for $\mathcal{L}_{\mathrm{Rep}}$ and the teacher-forcing inputs at R positions. At inference, the right panel is bypassed entirely: the decoder predicts representation tokens from the text prompt alone, and these tokens remain in context to guide pixel-space diffusion.}
\label{fig:overview}
\end{figure}

The design principle behind Representation Forcing is simple: in understanding, the encoder maps images to representations that capture high-level structure; in generation, the decoder mirrors this by predicting representations from text alone, before rendering them into pixels.
In Sec.~\ref{sec:3.1}, we describe where these representations come from and how they are formulated. In Sec.~\ref{sec:3.2}, we show how the decoder predicts them to guide pixel-space generation. In Sec.~\ref{sec:3.3}, we present the overall architecture and training objective. The full training pipeline is illustrated in Figure~\ref{fig:overview}.

\subsection{Representations from Understanding}
\label{sec:3.1}

Rather than relying on an external latent space, we seek an intermediate representation that captures high-level structure \emph{from within the model itself}, so that pixel-space diffusion can focus on low-level rendering. The understanding encoder provides a natural source: its features, trained for visual comprehension, already encode the structural content we need. The encoder is jointly trained with the rest of the model. To make its features predictable by the decoder under the same next-token objective as text, and easy to sample at inference, we discretize them into a sequence of \emph{visual representation tokens} via vector quantization. Beyond enabling unified training, discretization encourages the representations to retain high-level structure while discarding fine-grained details, which achieves the factorization we seek between representation prediction and pixel rendering. We validate this choice against continuous regression in Sec.~\ref{sec:ablation}.

We perform this discretization through online vector quantization, which requires no separate pretrained tokenizer. Since the encoder is jointly trained, its features evolve throughout training; we therefore extract features from an exponential moving average (EMA) copy of the encoder, providing slow-moving targets that keep the discrete assignments stable. Specifically, we use patch-level features from the last layer of the EMA encoder, before the final norm. 

We maintain a codebook of $K$ prototype embeddings, and tokenization proceeds in three steps. First, we compute the cosine similarity between each patch-level feature and all $K$ prototypes, yielding a score matrix over the batch. Second, rather than greedily assigning each feature to its most similar prototype---which would let a few popular prototypes absorb most features and collapse the codebook---we balance the score matrix with a Sinkhorn--Knopp normalization following SwAV~\cite{swav}, which softly constrains every prototype to receive an equal share of the batch. Third, each feature is assigned to its highest-scoring prototype under the balanced scores, producing a discrete token index: intuitively, a feature keeps its most similar prototype unless that prototype is over-subscribed, in which case it falls back to its next-best match. Finally, the codebook is updated online with EMA updates following VQ-VAE~\cite{vqvae}, where each prototype moves toward the mean of the features assigned to it. This yields one representation token per spatial patch, forming a sequence that mirrors the spatial layout of the image and can be predicted in raster-scan order.

% We maintain a codebook of $K$ learnable prototype embeddings; for each patch-level feature, we compute its cosine similarity to all prototypes and assign it to the nearest one, producing a discrete token index. The codebook is updated online via exponential moving average (EMA) following VQ-VAE~\cite{vqvae}, where each prototype moves toward the mean of the features assigned to it. To prevent codebook collapse, we balance the assignments with a Sinkhorn--Knopp normalization following SwAV~\cite{swav}. This yields one representation token per spatial patch, forming a sequence that mirrors the spatial layout of the image and can be predicted in raster-scan order.

\subsection{Generating Pixels via Predicted Representations}
\label{sec:3.2}

With the representation tokens defined, we now describe how the decoder uses them to generate images. During training, the EMA encoder provides the representation tokens as ground-truth targets, and the decoder learns to predict them autoregressively under cross-entropy loss, within the same next-token prediction stream as text. At inference, the encoder is no longer involved: the decoder produces the representation tokens from the text prompt alone. We call this process \textbf{Representation Forcing}: on one hand, the understanding encoder's representations force the decoder to learn the same high-level visual structure; on the other hand, the decoder's predicted representations force the pixel generation process to follow the intended semantic layout.

Once predicted, the representation tokens remain in the sequence and serve as in-context conditioning for pixel-space generation; the pixel patches form the final image, while the representation tokens themselves are not part of the visible output. The same backbone performs this generation through flow matching. Following JiT~\cite{jit}, we adopt x-prediction with velocity loss. Given clean patches $\mathbf{x}$ and Gaussian noise $\boldsymbol{\epsilon} \sim \mathcal{N}(\mathbf{0}, \mathbf{I})$, noisy patches at time $t \in [0, 1]$ are
\begin{equation}
    \mathbf{z}_t = t\,\mathbf{x} + (1 - t)\,\boldsymbol{\epsilon}.
\end{equation}
The decoder predicts $\mathbf{x}_\theta$, and the flow-matching loss is
\begin{equation}
    \mathcal{L}_{\text{FM}} = \mathbb{E} \big\| \mathbf{v}_\theta - \mathbf{v} \big\|^2,
\end{equation}
where $\mathbf{v} = \mathbf{x} - \boldsymbol{\epsilon}$ and $\mathbf{v}_\theta = (\mathbf{x}_\theta - \mathbf{z}_t)/(1 - t)$.

At the sequence level, as shown in Figure~\ref{fig:overview}, the model processes a unified token sequence structured as [text tokens, representation tokens, pixel patches]. Within the shared self-attention of the backbone, text and representation tokens follow causal attention as in standard autoregressive modeling, while the noisy pixel patches attend bidirectionally to each other and causally to all preceding text and representation tokens. This attention pattern is what makes the representation tokens act as in-context conditioning: the high-level structure they encode flows into pixel generation through standard self-attention, without any additional cross-attention or injection module.

\subsection{Training and Inference}
\label{sec:3.3}
Our model adopts the Mixture-of-Transformers (MoT) architecture following BAGEL~\cite{bagel}. All tokens share the same multi-head self-attention layers, but are routed to modality-specific feed-forward experts based on their type. We maintain three groups of experts: one for multimodal understanding, one for representation token prediction, and one for pixel generation. Each representation token is embedded as the sum of two learnable embeddings: a 2D spatial position embedding, and a token identity embedding indexed by the discrete token ID. The latter is stored in a separate $K$-entry table.

The model is trained end-to-end with a combined objective:
\begin{equation}
    \mathcal{L} = \mathcal{L}_{\text{LM}} + \mathcal{L}_{\text{FM}} + \mathcal{L}_{\text{Rep}},
\end{equation}
where $\mathcal{L}_{\text{LM}}$ is the cross-entropy loss for text next-token prediction, $\mathcal{L}_{\text{Rep}}$ is the cross-entropy loss for representation token prediction, and $\mathcal{L}_{\text{FM}}$ is the flow-matching loss for pixel generation following the x-prediction formulation in Sec.~\ref{sec:3.2}. To support classifier-free guidance at inference, we independently drop the text conditioning and the representation token sequence with probability 0.1 during training.

At inference, generation proceeds in two stages. The decoder first produces the full representation token sequence autoregressively from the text prompt. Conditioned on both the text and the predicted representation tokens, the decoder then performs iterative denoising from Gaussian noise to synthesize the final image directly in pixel space. Classifier-free guidance is applied to both conditions.

\section{Experiments}

\subsection{Experimental Setup}
\label{sec:exp_setup}

\begin{table}[t]
\centering
\small
\setlength{\tabcolsep}{2.8pt}
\caption{\textbf{Evaluation of text-to-image generation.} $\dagger$ refers to methods using LLM rewriter. Our pixel-space model with Representation Forcing (RF-Pixel) matches state-of-the-art VAE-based unified models on both GenEval and DPG-Bench, without relying on any pretrained VAE.}
\label{tab:geneval}
\begin{tabular}{l|ccccccc|c}
\toprule
& \multicolumn{7}{c|}{GenEval} & DPG \\
\cmidrule(lr){2-8} \cmidrule(lr){9-9}
Model & Single Obj. & Two Obj. & Counting & Colors & Position & Color Attri. & Overall$\uparrow$ & Overall$\uparrow$ \\
\midrule
\multicolumn{9}{l}{\textit{Generation Only}} \\
\midrule
PixArt-$\alpha$~\cite{pixart}     & 0.98 & 0.50 & 0.44 & 0.80 & 0.08 & 0.07 & 0.48 & 71.11 \\
SDv2.1~\cite{stable_diff}               & 0.98 & 0.51 & 0.44 & 0.85 & 0.07 & 0.17 & 0.50 & 68.09 \\
DALL-E 2~\cite{dalle2}            & 0.94 & 0.66 & 0.49 & 0.77 & 0.10 & 0.19 & 0.52 & -- \\
SDXL~\cite{sdxl}                  & 0.98 & 0.74 & 0.39 & 0.85 & 0.15 & 0.23 & 0.55 & 74.65 \\
DALL-E 3~\cite{dalle3}            & 0.96 & 0.87 & 0.47 & 0.83 & 0.43 & 0.45 & 0.67 & 83.50 \\
SD3-Medium~\cite{sd3}             & 0.99 & 0.94 & 0.72 & 0.89 & 0.33 & 0.60 & 0.74 & 84.08 \\
FLUX.1-dev$^\dagger$~\cite{flux}  & 0.98 & 0.93 & 0.75 & 0.93 & 0.68 & 0.65 & 0.82 & 84.00 \\
Seedream 3.0~\cite{seedream}      & 0.99 & 0.96 & 0.91 & 0.93 & 0.47 & 0.80 & 0.84 & 88.27 \\
Z-Image-Turbo~\cite{zimageturbo}  & 1.00 & 0.95 & 0.77 & 0.89 & 0.65 & 0.68 & 0.82 & 84.86 \\
Qwen-Image~\cite{qwenimage}       & 0.99 & 0.92 & 0.89 & 0.88 & 0.76 & 0.77 & 0.87 & 88.32 \\
\midrule
\multicolumn{9}{l}{\textit{Unified Models}} \\
\midrule
Chameleon~\cite{chameleon}         & --   & --   & --   & --   & --   & --   & 0.39 & -- \\
LWM~\cite{lwm}                    & 0.93 & 0.41 & 0.46 & 0.79 & 0.09 & 0.15 & 0.47 & -- \\
SEED-X~\cite{seedx}               & 0.97 & 0.58 & 0.26 & 0.80 & 0.19 & 0.14 & 0.49 & -- \\
TokenFlow-XL~\cite{tokenflow}     & 0.95 & 0.60 & 0.41 & 0.81 & 0.16 & 0.24 & 0.55 & 73.38 \\
ILLUME~\cite{illume}               & 0.99 & 0.86 & 0.45 & 0.71 & 0.39 & 0.28 & 0.61 & -- \\
Janus~\cite{janus}                & 0.97 & 0.68 & 0.30 & 0.84 & 0.46 & 0.42 & 0.61 & -- \\
Transfusion~\cite{transfusion}    & --   & --   & --   & --   & --   & --   & 0.63 & -- \\
Emu3$^\dagger$~\cite{emu3}        & 0.99 & 0.81 & 0.42 & 0.80 & 0.49 & 0.45 & 0.66 & 81.60 \\
Show-o~\cite{showo}               & 0.98 & 0.80 & 0.66 & 0.84 & 0.31 & 0.50 & 0.68 & -- \\
Show-o2~\cite{showo2}             & 1.00 & 0.87 & 0.58 & 0.92 & 0.52 & 0.62 & 0.76 & 86.14 \\
Janus-Pro-7B~\cite{janus-pro}     & 0.99 & 0.89 & 0.59 & 0.90 & 0.79 & 0.66 & 0.80 & 84.19 \\
MetaQuery-XL$^\dagger$~\cite{metaquery} & -- & -- & -- & -- & -- & -- & 0.80 & 82.05 \\
BLIP3-o~\cite{blip3o}             & --   & --   & --   & --   & --   & --   & 0.84 & 81.60 \\
UniWorld-V1$^\dagger$~\cite{uniworld} & 0.98 & 0.93 & 0.81 & 0.89 & 0.74 & 0.71 & 0.84 & 81.38 \\
OmniGen2$^\dagger$~\cite{omnigen2} & 0.99 & 0.96 & 0.74 & 0.98 & 0.71 & 0.75 & 0.86 & 83.57 \\
BAGEL~\cite{bagel}                & 0.99 & 0.94 & 0.81 & 0.88 & 0.64 & 0.63 & 0.82 & 85.07 \\
BAGEL$^\dagger$~\cite{bagel}      & 0.98 & 0.95 & 0.84 & 0.95 & 0.78 & 0.77 & 0.88 & -- \\
\midrule
\rowcolor{gray!10}\textbf{RF-Pixel (ours)} & 0.99 & 0.93 & 0.84 & 0.89 & 0.74 & 0.66 & 0.84 & 84.15 \\
\rowcolor{gray!10}\textbf{RF-Pixel (ours)}$^\dagger$ & 0.98 & 0.95 & 0.88 & 0.87 & 0.92 & 0.70 & 0.88 &-  \\
\bottomrule
\end{tabular}
\end{table}

\textbf{Architecture.}
Our model is initialized from Qwen3-A3B~\cite{qwen3}, a pretrained Mixture-of-Experts language model with 3B active parameters per token, and follows the Mixture-of-Transformers (MoT) architecture~\cite{bagel}: all tokens share self-attention layers but are routed to one of three modality-specific feed-forward expert pools---understanding, representation prediction, and pixel generation. The image encoder is DINOv3 ViT-H+/16~\cite{dinov3} with NaViT-style variable-resolution support~\cite{navit}, jointly trained with the rest of the model. We use a codebook of $K{=}16{,}384$ prototypes for the online vector quantization. For pixel-space generation we use a $16{\times}16$ patch size and adopt $x$-prediction with velocity loss~\cite{jit}. The pooling factor is a hyperparameter that trades off representation granularity against sequence length; we use $2{\times}2$ pooling throughout, yielding $N$ representation tokens for every $4N$ pixel patches over a shared spatial layout.

\textbf{Data.}
We follow the data construction and filtering pipeline of BAGEL~\cite{bagel}, training on a mixture of (1)~text-only data for language modeling and (2)~large-scale text--image pairs covering both image-to-text understanding (general VQA, document comprehension, spatial reasoning) and text-to-image generation.

\textbf{Training.}
We adopt a three-stage training strategy following ~\cite{bagel}: (i)~\emph{alignment}: with the backbone and encoder frozen, we train only the MLP connector for 10K iterations; (ii)~\emph{joint pre-training}: we unfreeze all components and jointly optimize on text and text--image pairs at resolutions up to 256 for 50K iterations; (iii)~\emph{continued training}: we extend resolutions up to 1024 for 20K iterations. Throughout training, image resolutions are sampled dynamically within the per-stage maximum and packed via NaViT-style variable-resolution batching.  More details are in the appendix.

\textbf{Baselines.}
For controlled comparison, we train VAE-based variants using the WanX-2.1 VAE~\cite{wanx}, replacing the pixel input/output with VAE latents while keeping the rest of the architecture, training data, and optimization identical. The four variants in our controlled study (Pixel, Pixel+RF, VAE, VAE+RF) differ \emph{only} in (a)~the generation pathway and (b)~whether RF is enabled. Ablations are conducted at 256 resolution; main results are reported at 1024 resolution.

\subsection{Image Generation}
\label{sec:t2i}

We evaluate our pixel-space model with RF (denoted RF-Pixel in Table~\ref{tab:geneval}) on two standard text-to-image benchmarks: GenEval~\cite{geneval} for compositional generation and DPG-Bench~\cite{dpg} for dense prompt following. We compare against both generation-only models and unified multimodal models. Existing unified models all rely on separately pretrained generative components—either a frozen VAE/VQVAE within a single backbone (e.g., BAGEL, Show-o, Janus-Pro), or an external diffusion module conditioned on LLM-predicted features (e.g., BLIP3-o, MetaQuery, SEED-X). In contrast, our model operates entirely in pixel space without any separately pretrained generative module.

As shown in Table~\ref{tab:geneval}, without LLM rewriter, RF-Pixel achieves a GenEval overall score of 0.84, slightly outperforming the BAGEL baseline (0.82) and matching unified models such as BLIP3-o (0.84). With LLM rewriter, RF-Pixel reaches 0.88, matching the state-of-the-art among unified models. On DPG-Bench, RF-Pixel scores 84.15 without rewriter, comparable to state-of-the-art VAE-based unified models. These results demonstrate that Representation Forcing effectively closes the quality gap between pixel-space and VAE-based generation, enabling an end-to-end pixel-space unified model to match VAE-based counterparts.

\subsection{Image Understanding}
\label{sec:understanding}
\begin{table}[t]
\centering
\footnotesize
\setlength{\tabcolsep}{5pt}
\caption{\textbf{Impact of RF on understanding.} We compare understanding performance with and without RF under both VAE-based and pixel-space generation. MME$^*$ reports the average accuracy across all perception and cognition questions. {\tiny\color{teal}+x}/{\tiny\color{red}$-$x} denotes the change from adding RF. RF improves both settings, with large gains on general visual understanding and only slight reductions on document-oriented tasks. Pixel+RF benefits more from RF than VAE+RF (6/8 vs.\ 5/8 benchmarks improved), and outperforms VAE+RF on 6 out of 8 benchmarks.}
\label{tab:understanding}
\begin{tabular}{l|ccccc|ccc}
\toprule
& \multicolumn{5}{c|}{General Visual Understanding} & \multicolumn{3}{c}{Document \& Diagram} \\
\cmidrule(lr){2-6} \cmidrule(lr){7-9}
& MMMU & HalluBench & MME$^*$ & BLINK & RealWorldQA & AI2D & DocVQA & ChartQA \\
\midrule
VLM-only & 56.2 & 65.0 & 79.7 & 56.2 & 65.8 & 90.3 & 89.3 & 86.0 \\
\midrule
VAE    & 51.0 & 55.7 & 71.3 & 52.2 & 65.2 & 90.7 & 90.0 & 78.8 \\
VAE+RF & 49.6{\tiny\color{red}\,$-$1.4} & 61.3{\tiny\color{teal}\,+5.6} & 79.3{\tiny\color{teal}\,+8.0} & 52.9{\tiny\color{teal}\,+0.7} & 66.6{\tiny\color{teal}\,+1.4} & 87.8{\tiny\color{red}\,$-$2.9} & 88.3{\tiny\color{red}\,$-$1.7} & 80.5{\tiny\color{teal}\,+1.7} \\
\midrule
Pixel    & 49.9 & 63.7 & 76.6 & 49.4 & 63.1 & 85.8 & 90.0 & 81.7 \\
Pixel+RF & 54.2{\tiny\color{teal}\,+4.3} & 64.8{\tiny\color{teal}\,+1.1} & 80.2{\tiny\color{teal}\,+3.6} & 53.0{\tiny\color{teal}\,+3.6} & 65.8{\tiny\color{teal}\,+2.7} & 90.3{\tiny\color{teal}\,+4.5} & 88.0{\tiny\color{red}\,$-$2.0} & 81.3{\tiny\color{red}\,$-$0.4} \\
\bottomrule
\end{tabular}
\end{table}

Beyond generation quality, we compare how different generation pathways affect understanding performance. As shown in Table~\ref{tab:understanding}, we train four generation variants---VAE-based and pixel-space, each with and without RF---on top of the same first-stage VLM baseline, using identical architecture and training data. Since we focus on pretraining-stage comparison, no post-training is applied. We include the VLM-only baseline as a reference.

We evaluate on 8 benchmarks spanning two categories: (1)~general visual understanding, including MMMU~\cite{mmmu}, HallusionBench~\cite{hallubench}, MME~\cite{mme}, BLINK~\cite{blink}, and RealWorldQA~\cite{rwqa}, which test general visual understanding, hallucination robustness, and real-world perception; and (2)~document and diagram understanding, including AI2D~\cite{ai2d}, DocVQA~\cite{docvqa}, and ChartQA~\cite{chartqa}, which test structured visual comprehension.

RF consistently improves understanding under both generation pathways. For pixel-space generation, RF improves 6 of 8 benchmarks, with substantial gains on MMMU~($+4.3$), MME~($+3.6$), BLINK~($+3.6$), AI2D~($+4.5$), and RealWorldQA~($+2.7$), with small reductions on DocVQA~($-2.0$) and ChartQA~($-0.4$). For VAE-based generation, RF improves 5 of 8 benchmarks, notably HalluBench~($+5.6$) and MME~($+8.0$). The improvements are concentrated on benchmarks that test high-level visual understanding---object recognition, spatial comprehension, and scene-level perception---which aligns with the nature of the representation tokens: they are derived from the understanding encoder and encode semantic structure rather than fine-grained details. The reductions on DocVQA and ChartQA, which rely heavily on precise text recognition and layout parsing, suggest that these capabilities are less directly supported by representation-level guidance.

Pixel+RF outperforms VAE+RF on 6 out of 8 benchmarks. We attribute this to the removal of the external VAE latent space, which allows understanding and generation to share a single representation space more tightly. This result aligns with our broader motivation of moving toward bottleneck-free unified multimodal models.

\subsection{Ablation Studies}
\label{sec:ablation}

We conduct ablation studies to analyze the key design choices of Representation Forcing in Table~\ref{tab:ablation}.

\begin{table}[t]
\centering
% \vspace{-15pt}
\caption{\textbf{Ablation studies.} All experiments are conducted under pixel-space generation at 256 resolution unless otherwise noted. (a)~Effect of RF under both pixel-space and VAE-based generation. (b)~Comparison between RF (decoder prediction) and REPA (auxiliary alignment) as representation guidance strategies. (c)~Discrete vs.\ continuous representation token formulation. (d)~Effect of codebook size $K$. (e)~Choice of understanding encoder evaluated on VLM-only benchmarks.}
\label{tab:ablation}
\begin{subtable}[t]{0.32\textwidth}
\centering
\small
\caption{\textbf{RF on pixel-space and VAE-based generation.}}
\label{tab:abl_representations}
\begin{tabular}{lcc}
\toprule
 & Pixel & VAE \\
\midrule
w/o & 0.25 & 0.52 \\
w/  & \textbf{0.76} & \textbf{0.77} \\
\bottomrule
\end{tabular}
\end{subtable}
\hfill
\begin{subtable}[t]{0.32\textwidth}
\centering
\small
\caption{\textbf{Prediction vs.\ alignment.}}
\label{tab:abl_integration}
\begin{tabular}{lc}
\toprule
 & GenEval$\uparrow$ \\
\midrule
w/o      & 0.25 \\
REPA     & 0.43 \\
RF (ours) & \textbf{0.76} \\
\bottomrule
\end{tabular}
\end{subtable}
\hfill
\begin{subtable}[t]{0.32\textwidth}
\centering
\small
\caption{\textbf{Rep.\ token formulation.}}
\label{tab:abl_target}
\begin{tabular}{lc}
\toprule
 & GenEval$\uparrow$ \\
\midrule
w/o        & 0.25 \\
Continuous & 0.26 \\
Discrete   & \textbf{0.76} \\
\bottomrule
\end{tabular}
\end{subtable}

\noindent
\begin{subtable}[t]{0.32\textwidth}
\centering
\small
\caption{\textbf{Rep.\ token codebook size.}}
\label{tab:abl_codebook}
\begin{tabular}{lc}
\toprule
 & GenEval$\uparrow$ \\
\midrule
$K$=16384 & 0.76 \\
$K$=32768 & \textbf{0.77} \\
\bottomrule
\end{tabular}
\end{subtable}
\hfill
\begin{subtable}[t]{0.65\textwidth}
\small
\setlength{\tabcolsep}{3pt}
\caption{\textbf{Understanding encoder.}}
\label{tab:abl_encoder}
\begin{tabular}{lccccc}
\toprule
 & MMMU & HalluBench & MME$^*$ & BLINK & RealWorldQA \\
\midrule
SigLIP2 & \textbf{56.3} & 64.4 & 76.8 & 53.4 & 62.5 \\
DINOv3 & 56.2 & \textbf{65.0} & \textbf{79.7} & \textbf{56.2} & \textbf{65.8} \\
\bottomrule
\end{tabular}
\end{subtable}
\end{table}

\begin{figure}[t]
\centering
\includegraphics[width=\linewidth]{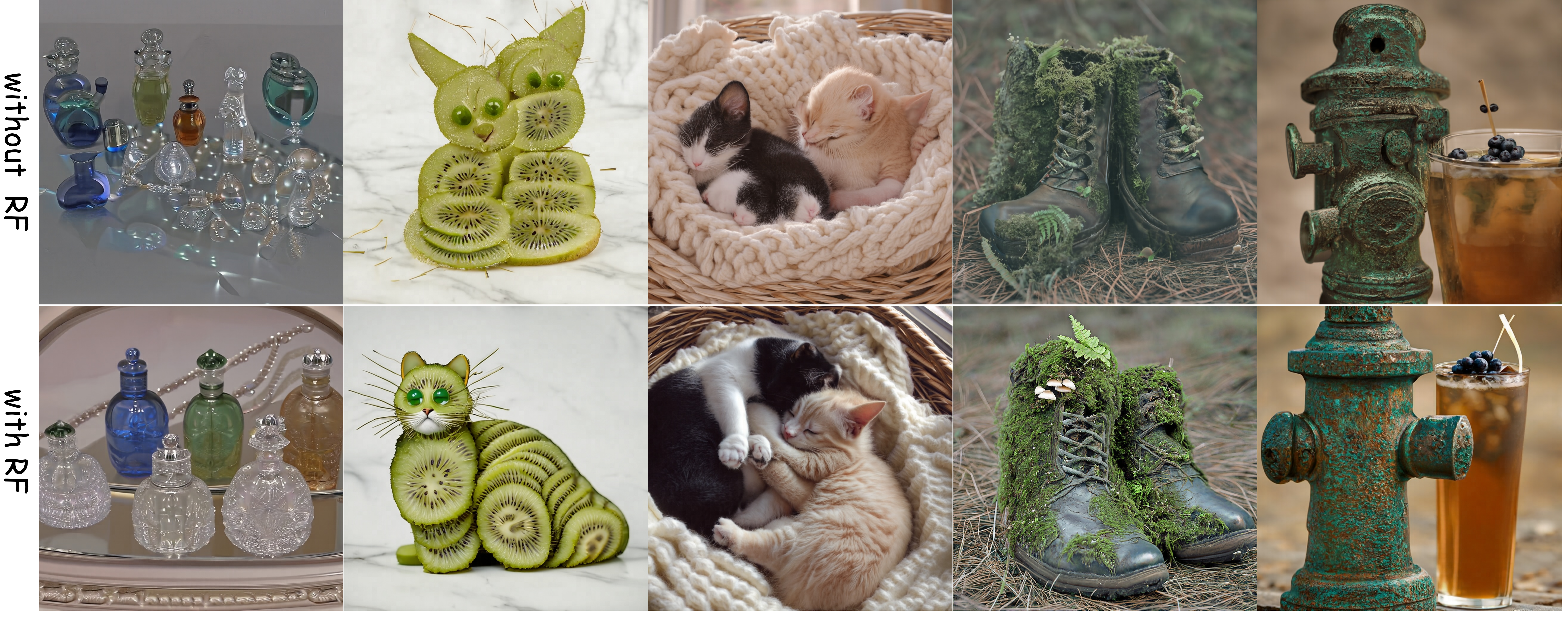}
\caption{\textbf{Qualitative comparison of pixel-space generation with and without RF.} Without RF, the model tends to produce images with poor structure, such as distorted object shapes and incoherent compositions. With RF, the model generates more coherent structures by first predicting high-level visual representations before pixel rendering, which provides explicit structural guidance for the diffusion process.}
\label{fig:compare}
\end{figure}

\textbf{Representations are critical for pixel-space generation.}
We evaluate RF under both pixel-space and VAE-based generation, with results shown in Table~\ref{tab:abl_representations}. All four variants share the same architecture and training setup, differing only in the generation pathway and the presence of RF. Without RF, pixel-space generation scores only 0.25 on GenEval, while VAE-based generation reaches 0.52. As illustrated in Figure~\ref{fig:compare}, the pixel-space model without RF tends to produce images with poor structure, such as distorted objects and incoherent compositions, suggesting that the model struggles to jointly learn high-level layout and low-level detail from raw pixels alone. With RF, pixel-space generation jumps to 0.76, matching the VAE-based counterpart at 0.77, and the generated images become structurally coherent. RF also improves VAE-based generation from 0.52 to 0.77, confirming that explicit representation guidance benefits both settings, though the effect is most pronounced in pixel space.

\textbf{Decoder prediction outperforms auxiliary alignment.}
RF is not the only way to incorporate visual representations into
generation. REPA~\cite{repa} takes an alternative approach by
adding an auxiliary loss on the diffusion side that encourages the
model's intermediate features to align with encoder representations.
We apply both methods to the same pixel-space UMM under identical
training conditions, using the same visual encoder (DINOv3~\cite{dinov3}) as the
representation source, with results shown in
Table~\ref{tab:abl_integration}. Without any guidance, pixel-space
generation scores only 0.25. REPA improves this to 0.43, confirming
that representation guidance helps, but the gain remains limited.
RF achieves 0.76, substantially outperforming REPA. We attribute
this difference to where the representation enters the generation
process. REPA encourages feature similarity as a side objective,
but the aligned features do not explicitly condition the generation
output during inference. RF places predicted representations directly in the
decoder's sequence, where pixel patches attend to them through
shared self-attention. In high-dimensional pixel space, this direct
conditioning proves more effective than implicit feature
alignment.

\textbf{Discrete prediction outperforms continuous regression.}
The representation tokens can be formulated in different ways. Beyond our discrete approach with autoregressive cross-entropy prediction, an alternative is to predict continuous features directly, for example by adding a diffusion head at each representation position to causally regress the encoder features. We compare both formulations under the same setting (Table~\ref{tab:abl_target}). Continuous regression scores only 0.26, providing no improvement over the baseline without RF, while discrete tokens achieve 0.76. We attribute this to two factors. First, causally predicting high-dimensional continuous vectors is prone to error accumulation, where small prediction errors at early positions compound and degrade later predictions. Discrete tokens avoid this by reducing each position to a categorical choice, which is more robust under sequential prediction. Second, discretization naturally encourages the representations to retain high-level structure while discarding fine-grained detail---precisely the factorization that RF is designed to provide. Continuous targets preserve more low-level information, undermining this factorization.

\textbf{Codebook size.} The codebook size $K$ controls the expressiveness of the discrete representation space. We compare $K$=16,384 and $K$=32,768 in Table~\ref{tab:abl_codebook}. The two settings perform comparably (0.76 vs.\ 0.77), suggesting that the model is not sensitive to codebook size within this range. We use $K$=16,384 in all other experiments, which is consistent with common practice in vector quantization.

\textbf{Understanding encoder.} The choice of understanding encoder determines the quality of the representations that RF learns to predict. We compare SigLIP2~\cite{tschannen2025siglip2}, a contrastive vision-language model, and DINOv3~\cite{dinov3}, a self-supervised vision model, as the image encoder backbone (Table~\ref{tab:abl_encoder}). We find DINOv3 outperforms SigLIP2 on 4 out of 5 understanding benchmarks in our setting. We attribute DINOv3's advantage to its self-supervised training objective, which produces features with richer spatial and structural information compared to SigLIP2's language-aligned features. This aligns with RF's design: the representation tokens are meant to capture visual structure, which benefits from an encoder that prioritizes spatial fidelity over text alignment. We use DINOv3 as the default encoder.

\section{Discussion}
\label{sec:discussion}

\textbf{Limitations.} Due to computational constraints, our model is initialized from a pretrained large language model rather than trained from scratch on multimodal data. While this provides a strong language-grounded starting point, fully from-scratch multimodal pretraining may yield richer joint representations and is an important direction for future work. We also focus on still-image generation and do not extend RF to video or other temporal modalities.

\textbf{Conclusion.}
In this paper, we present Representation Forcing, a simple method for pixel-space image generation in unified multimodal models. The key idea is to let the decoder predict its own understanding representations as autoregressive targets before rendering pixels, providing structural guidance to pixel-space diffusion from within the same sequence. Our experiments show that this single mechanism is enough to close the quality gap with VAE-based generation while also improving multimodal understanding. The same representation serves both directions---interpreting visual inputs and guiding their generation---pointing to a closer integration of perception and generation. We see RF as a step toward fully end-to-end native multimodal learning, where all multimodal capabilities are acquired directly from raw inputs within a single model. We hope this work inspires further research in this direction.

\clearpage

\bibliographystyle{plainnat}
\bibliography{main}

\clearpage

\beginappendix

\section{Implementation Details}

\textbf{Training.} We train using AdamW~\cite{adamw} ($\beta_1{=}0.9$, $\beta_2{=}0.95$, $\epsilon{=}10^{-8}$, weight decay $0.1$, gradient clipping $1.0$). The learning rate follows linear warmup followed by a constant schedule, with a base rate of $5{\times}10^{-5}$ in Stages 1--2 and $2.5{\times}10^{-5}$ in Stage 3 for high-resolution stability. Newly initialized generation-related parameters use a $4{\times}$ multiplier on the base rate, while the LLM backbone keeps the base rate. Each GPU processes sequences of $32{,}768$ tokens, packed via NaViT-style variable-resolution batching. The online vector quantization codebook is updated via momentum (decay $0.9999$, $1$ Sinkhorn-Knopp iteration, temperature $0.5$); pseudocode is provided in Algorithm~\ref{alg:online_vq}. For classifier-free guidance, we independently drop the text condition and the entire representation token sequence, each with probability $0.1$.

\textbf{Inference.} We maintain an exponential moving average of model parameters with decay $0.9999$ and perform inference using the EMA model. Generation proceeds in two stages: the decoder first produces the full representation token sequence autoregressively from the text prompt using top-$k$ sampling, then denoises Gaussian noise into pixel patches over $25$ flow-matching steps with dynamic timestep shifting~\cite{sd3}, conditioned on the text and predicted representation tokens. We apply two-condition CFG with $w_{\text{rep}}{=}2.0$ for representation token sampling and $w_{\text{pix}}{=}3.0$ for pixel patch denoising.

\section{Online Vector Quantization Algorithm}
\label{sec:appendix-vq}

\begin{algorithm}[h]
\caption{Pseudocode of Online Vector Quantization (PyTorch-like).}
\label{alg:online_vq}
\begin{lstlisting}[language=Python]
# f_m: EMA understanding encoder
# X: batch of samples (BxHxWx3)
# C: visual prototypes (KxD)
# m: momentum (default 0.9999)
# t: temperature (default 0.5)

Z = f_m(X).view(B * L, D)           # extract continuous features via EMA encoder
Z = normalize(Z, dim=1)
score = matmul(Z, C.T) / t          # pairwise cosine sims: (B*L, K)
score = softmax(score, dim=1)       # row-normalize
score = softmax(score, dim=0)       # column-normalize (Sinkhorn-Knopp, 1 iteration)
A = argmax(score, dim=1)            # discrete assignment: (B*L,)

N_k, C_new = zeros(K), zeros(K, D)
A_c = A.view(B * L, 1).expand(B * L, K)
C_new = scatter_add(C_new, dim=0, index=A_c, src=Z)
N_k = scatter_add(N_k, dim=0, index=A, src=ones(B * L))
C_new = normalize(C_new / N_k, dim=1) # new prototypes from assignments
C = m * C + (1 - m) * C_new           # momentum update
C = normalize(C, dim=1)
\end{lstlisting}
\end{algorithm}

\section{Broader Impact}

Like other text-to-image generation systems, Representation Forcing could potentially be misused to generate misleading or harmful visual content, including disinformation, non-consensual imagery, or deepfakes. Standard safeguards used for unified multimodal models---including safety filters, output watermarking, and controlled access---apply to RF-based systems.

\end{document}